\documentclass[runningheads]{llncs}
\usepackage{graphicx}
\usepackage{amsmath}
\usepackage{amssymb}
\usepackage{csquotes}

\usepackage{caption}
\usepackage{subcaption}

\usepackage{multirow}
\usepackage{cleveref}
\usepackage{adjustbox}
\usepackage[nolist,nohyperlinks]{acronym}
\begin{acronym}
\acro{VAE}{variational autoencoder}
\acro{DDPM}{Denoising Diffusion Probabilistic Model}
\acro{GAN}{generative adversarial network}
\acro{CNN}{convolutional neural network}
\acro{MRE}{mean radial error}
\acro{SDR}{success detection rate}
\acro{GT}{ground truth}
\end{acronym}

\def\etal{\emph{et al}}

\begin{document}
\title{Salt \& Pepper Heatmaps:\\
Diffusion-informed Landmark Detection Strategy}
\titlerunning{Salt \& Pepper Heatmaps}
\institute{Anonymous}
%
%

%

\author{Julian Wyatt \and Irina Voiculescu} 
\institute{Department of Computer Science, University of Oxford\\
\email{\{name\}.\{surname\}@cs.ox.ac.uk}}

\maketitle              

\begin{abstract}

%

Anatomical Landmark Detection is the process of identifying key areas of an image for clinical measurements. 
Each landmark is a single \ac{GT} point labelled by a clinician. A machine learning model predicts the locus of a landmark as a probability region represented by a heatmap. 
Diffusion models have increased in popularity for generative modelling due to their high quality sampling and mode coverage, leading to their adoption in medical image processing for semantic segmentation.
Diffusion modelling can be further adapted to learn a distribution over landmarks.
The stochastic nature of diffusion models captures fluctuations in the landmark prediction, which we leverage by blurring into meaningful probability regions.
In this paper, we reformulate automatic Anatomical Landmark Detection as a precise generative modelling task, producing a few-hot pixel heatmap. Our method achieves state-of-the-art MRE and comparable SDR performance with existing work. {\em Code will be made available upon publication.}

\end{abstract}

\begin{figure}
    \centering
         \includegraphics[width=.45\textwidth]{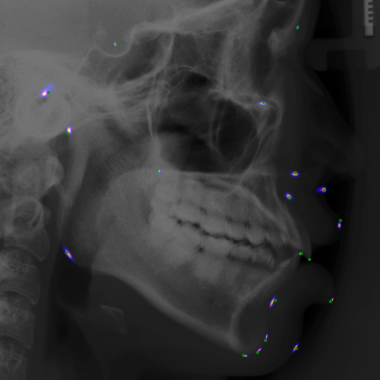}
                 \caption{Landmark heatmaps: hottest point avg. distance 0.72mm from green \ac{GT}.}
\end{figure}

\section{Introduction}


Landmark or keypoint detection is the process of highlighting a pixel and its close neighbourhood for use in measurement tasks. Traditional methods for landmark detection often rely on manual annotation, which is time-consuming and prone to large inter-annotator variance. The automation of this process in the medical domain improves the annotation speed and accuracy whilst allowing for annotator variance. Landmark detection is used in both diagnosis and treatment planning of various medical impairments such as broken~\cite{negrillo2020automatic} or dislocated bones~\cite{mccouat2021automatically}, or dental growth abnormalities~\cite{wang2016benchmark}. Consequently the refinement of landmark detection to within human error is beneficial for these tasks. In recent years, the application of deep learning to this automation has profoundly increased due to its accessibility, speed and accuracy. Most commonly, researchers utilise regressive approaches using \acp{CNN}~\cite{zhong2019attention}. For example, a heatmap regression outputs an image probability heatmap over the most likely landmark coordinates. To generate such heatmaps, we apply generative models.

One such architecture, \acfp{DDPM}~\cite{ho2020denoising} have continued to increase in popularity and traction in comparison to alternative generative models such as \acp{GAN}~\cite{goodfellow2014generative} and \acp{VAE}~\cite{kingma2013auto}. This is due to \acp{DDPM}' remarkable sample quality and mode coverage~\cite{xiao2022tackling} when generating images. A \ac{DDPM} generates data through a stochastic denoising procedure which learns gradually to transform a standard Gaussian distribution $\mathcal{N}(0,\mathbf{I})$ over $T$ steps onto the training data distribution. This introduces a sequential stochasticity and enables it to learn an uncertainty prediction over the training distribution, for example, using ensemble models~\cite{wolleb2022diffusion} and out-of-distribution detection~\cite{wyatt2022anoddpm}. However, one of the major drawbacks of \Acp{DDPM} is the slow sample time, with time complexity $\mathcal{O}(T)$ from $T$ forward passes through the neural network during sampling. Instead, we propose a single-step diffusion model for rapid analysis with a precise multi-step backbone for landmark detection. 

%

\section{Related Work}

Recent advancements of Anatomical Landmark Detection utilise variants of \Acp{CNN} to optimise a heatmap of probable landmark positions. The most popular variant, the U-Net~\cite{ronneberger2015u}, uses an encoder-decoder architecture with skip connections to pass features from the encoder to the decoder. Nearly 10 years later, U-Nets are still incredibly popular due to their strong, stable performance across most imaging domains such as image generation~\cite{ho2020denoising}, segmentation~\cite{ronneberger2015u}, and landmark detection~\cite{mccouat2022contour}. Moreover, the highest performing models for landmark detection typically adapt the U-Net formulation and utilise multi-resolution attentive methods such as Zhong \etal.~\cite{zhong2019attention} who combine separate coarse and local-level models for an overall better prediction. Additionally, these high-performing architectures utilise bespoke attention modules~\cite{zhong2019attention,ye2023uncertainty} which combine and attend to multi-level features through the encoding process. 

Most commonly, the heatmap is generated by optimising the negative log likelihood~\cite{mccouat2022contour}, while alternative approaches consider predicting displacement vectors via coordinate regression~\cite{wu2019facial}, or pixel regression using offset maps~\cite{chen2019cephalometric} to improve the overall radial error. Optimising solely coordinate regression generally struggles to learn high precision performance over image-space methods as coordinates may seem arbitrary to a model following a compressed image representation. However, a combination of heatmap and coordinate regression performs well~\cite{takahashi2023cephalometric}. 


\subsection{\Acfp{DDPM}}

The sub-field of generative modelling -- producing images using deep learning architectures -- has grown significantly over the last few years due to the impressive images from DALL-E 3~\cite{betker2023improving} and Cascaded diffusion models~\cite{ho2022cascaded}. This quality is enabled by \acp{DDPM}, a modern class of latent generative models. Unlike alternative classes of generative models, the latent variable has the same dimensionality as the original data, contributing to the improvement of sample quality. 

The \Ac{DDPM}~\cite{sohl2015deep,ho2020denoising} is defined as an iterative generative procedure that generates images by sampling from a Gaussian distribution $x_T \sim \mathcal{N}(0,\mathbf{I})$ and learns gradually to remove noise over $t=T,...,0$ steps until recovering an image from the training data distribution $q(x_0)$. This can be considered as a non-homogenous Markov chain where the previous image conditions the generation of the following image $p_\theta(x_{t-1}|x_t)$ until reaching the fully denoised sample $x_0$. The development of this model is split into two phases, the forward (training) phase and the reverse (inference) phase. 

The forward phase is defined as a single-step transition density function: \begin{equation}
    q(x_t|x_{t-1}) = \mathcal{N}(x_t| x_{t-1}\sqrt{1-\beta_t},\beta_{t}\mathbf{I}).
\end{equation}
Here, noise is gradually injected into the model utilising the variance schedule $\beta \in (0,1)$ where $\beta$ is formulated as a linear increase between $\beta_1=10^{-4}$ and $\beta_T = 0.02$ to allow $x_T$ to approximate an isotropic Gaussian distribution as $T\to\infty$ and $\beta_t\to0$~\cite{sohl2015deep}. Due to a reparameterisation from Ho \etal.~\cite{ho2020denoising}, this can be reformulated to generate $q(x_t | x_0)$ without the intermediate steps $x_{t-1},\ldots,x_1$ thus allowing fast sampling of $x_t$ for arbitrary $t$. By rewriting  $\alpha_t=1-\beta_t$, and $\bar{\alpha}_t = \prod^T_{i=0} \alpha_i$, we have:
\begin{align}
q(x_t | x_0) = \; & \mathcal{N}(x_t| x_0 \sqrt{\bar{\alpha}_t},(1-\bar{\alpha}_t)\mathbf{I}),\\ 
x_t = x_{0}\sqrt{\bar{\alpha}_t} +\;& \epsilon_{t}\sqrt{1 - \bar{\alpha}_t},  \quad \epsilon_t \sim \mathcal{N}(0,\mathbf{I}). \label{eq:q_sample}
\end{align}

The learned reverse process, parameterised by $\theta$, generates new samples by first sampling $x_T \sim \mathcal{N}(0,\mathbf{I})$ and iteratively generates a new sample according to
\begin{align}
    p_\theta (x_{t-1}|x_t) &= \mathcal{N} (x_{t-1}|\mu_\theta(x_t,t), \tilde{\beta}_t \mathbf{I})\\
    x_{t-1} &=  \mu_\theta(x_t,t) + \tilde{\beta}_t z, \quad z \sim \mathcal{N}(0,\mathbf{I})
\end{align} 
for $t = T, \ldots, 1$, and $\tilde{\beta}_t = \frac{1 - \bar{\alpha}_{t-1}}{1- \bar{\alpha}_t} \beta_t$. Where $\mu_\theta$ is an approximation of the mean training data distribution conditioned on the current sample and timestep, and is typically optimised using a deep U-Net~\cite{ronneberger2015u}. Parameterising \begin{equation} \label{eq:mu_original}
    \mu_\theta(x_t,t) = \frac{1}{\sqrt{\alpha_t}} \left(x_t - \frac{\beta_t}{\sqrt{1 - \bar{\alpha_t}}} \epsilon_\theta(x_t, t) \right),
\end{equation} we can estimate the injected noise as $\epsilon_\theta$, giving the following training objective:
\begin{equation}
    \mathcal{L}_{\text{simple}} = \mathbb{E}_{t \sim [1-T],x_0\sim q(x_0),\epsilon \sim \mathcal{N}(0,\mathbf{I})} [\,||\epsilon -\epsilon_\theta (x_t,t)||^2\,].
\end{equation}

\begin{figure}[t!]
    \centering
    \includegraphics[width=\linewidth]{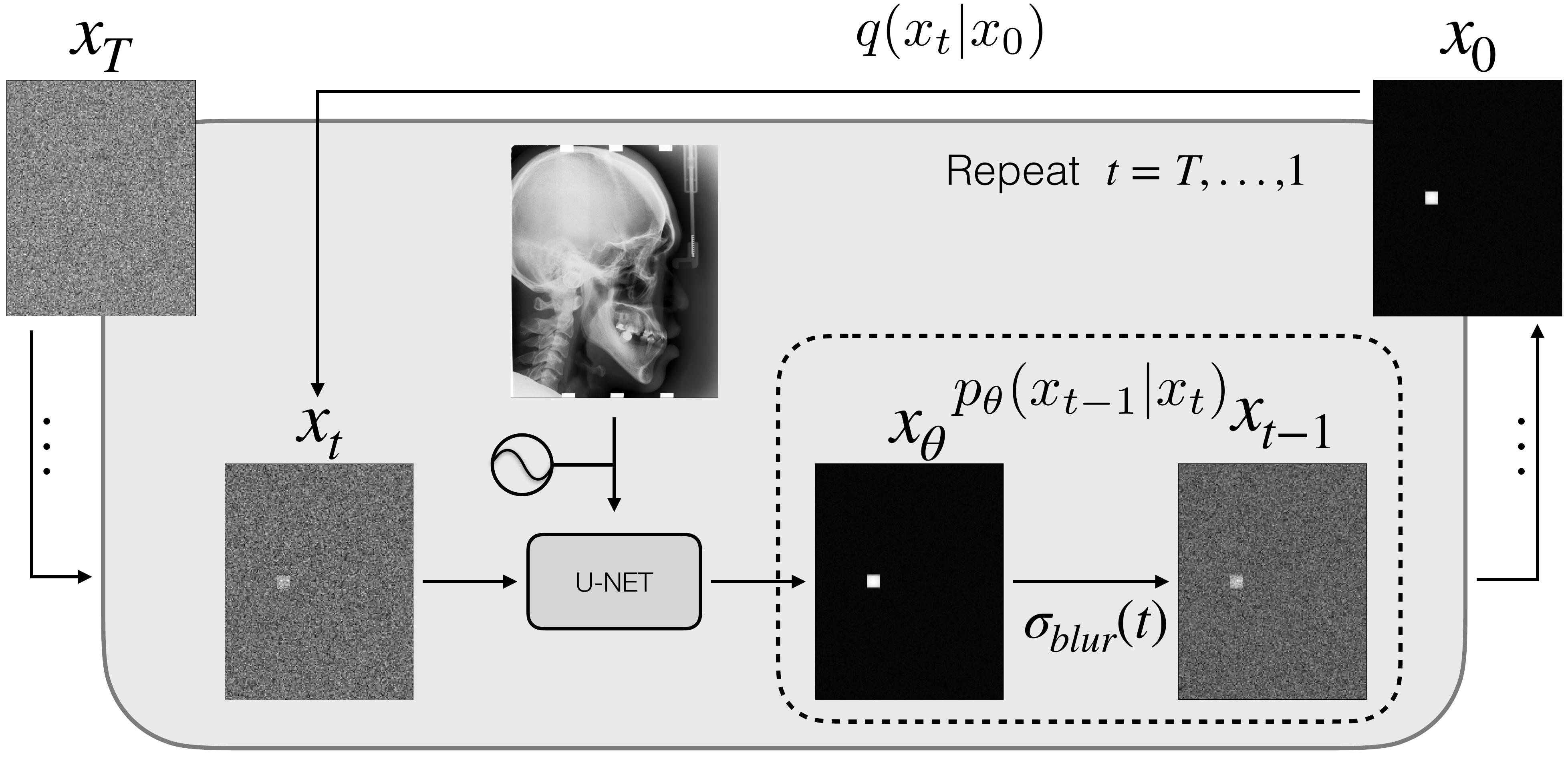}
    \caption{Reverse heatmap generation procedure: Initially using Gaussian noise at $x_T$; noise incrementally removed through the Markov Chain from $t=T,...,0$. 
    The parameterised model $x_\theta$ takes the positional encoding and reference image as context to build the formulation of the noise reduction approximation $p_\theta(x_{t-1}|x_t)$.
     (Exaggerated landmark size; only one of the $\mathbf{N}$ channels shown.)}
    \label{fig:Inference}
\end{figure}

\section{Methodology}

Iterative data generation through diffusion has been shown to improve the qualitative and quantitative quality of the generated data~\cite{dhariwal2021diffusion}. 
We formulate the landmark detection problem similarly to McCouat \etal.~\cite{mccouat2022contour} and aim to generate an $\mathbf{N}$ channel image, where $\mathbf{N}$ is the number of landmarks and each channel outputs a probability distribution over the most probable landmark positions for a given input raw image. To adapt the \ac{DDPM} work for landmark detection, we apply a similar approach used by Wolleb \etal.~\cite{wolleb2022diffusion} for segmentation: for a reference image $y$ with dimensions $(c,h,w)$, and ground truths $x$ with dimensions $(\mathbf{N},h,w)$, we train a \ac{DDPM} over the channel-wise concatenation $\mathbf{X}=y\oplus x$. However, at test-time, we require the reference image as a corresponding reference, therefore we apply \cref{eq:q_sample} on only $x$ to generate $x_t$ for some $t \in[0,T]$.

Furthermore, an optimal parameterisation distinguishes a single pixel over a noisy sample; in practice this turns out to be incredibly difficult, even for a well-trained model as a landmark can be considered as noise. Therefore, we revisit the derivation by Ho \etal.~\cite{ho2020denoising} and parameterise the model with the alternative setting: \begin{equation} \label{eq:mu_updated}
    \mu_\theta(x_t,t) = \frac{\sqrt{\alpha_t}(1-\bar{\alpha}_{t-1})}{1-\bar{\alpha}_t}x_t + \frac{\sqrt{\bar{\alpha}_{t-1}}\beta_t}{1-\bar{\alpha}_t}x_0.
\end{equation}
Thus, we predict the initial ground truth $x_{\theta} (x_t,t)$ as a single-step, updating the loss function to 
\begin{equation}
    \mathcal{L}_{s} = \mathbb{E}_{t \sim [1-T],x_0\sim q(x_0),\epsilon \sim \mathcal{N}(0,\mathbf{I})} [\,||x_0 -x_{\theta} (x_t,t)||^2\,].
\end{equation}
As a result, in the domain of landmark detection, this is much easier for a model to learn, resulting in drastically quicker training times over training $\epsilon_\theta$. Additionally, we want to learn a probability distribution, hence we also tweak the formulation to apply a spatial softmax $\sigma$ over the image dimensions and primarily optimise using a cross entropy loss: \begin{equation} \label{eq:nll-loss}
    \mathcal{L}_{nll} = \lambda_{nll} \cdot \mathbb{E}_{t \sim [1-T],x_0\sim q(x_0),\epsilon \sim \mathcal{N}(0,\mathbf{I})} [-x_0 \cdot \log(\sigma(x_{\theta} (x_t,t))+10^{-9})]
\end{equation}
This gives the following overall objective: 
\begin{equation}\label{eq:loss}
    \mathcal{L} = \lambda_{s} \cdot \mathcal{L}_{s} + \lambda_{nll} \cdot \mathcal{L}_{nll}.
\end{equation} Where we set $\lambda_{s}=0.01$ to reduce over-confident predictions and $\lambda_{nll}=1$. Note diffusion models require the scaling of the input data to match the same scale of the injected noise, therefore only when calculating $\mathcal{L}_{nll}$, we renormalise the data such that $x_0 \in [0,1]$.

\begin{figure}[t]
    \centering
     \begin{subfigure}{0.32\textwidth}
         \centering
         \includegraphics[width=\textwidth]{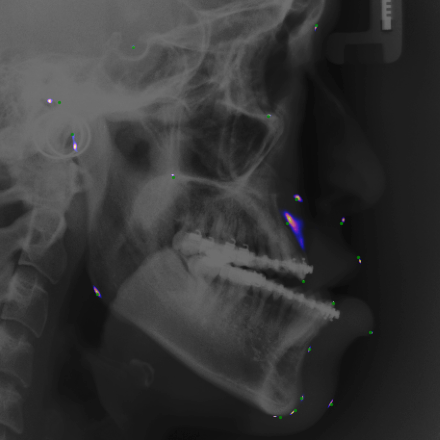}
         \caption{0.99mm MRE}
         \label{fig:good_sample}
     \end{subfigure}
     \hfill
     \begin{subfigure}{0.32\textwidth}
         \centering
         \includegraphics[width=\textwidth]{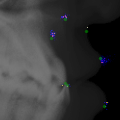}
         \caption{$x_\theta(x_T,T)$}
         \label{fig:zoomed_x0}
     \end{subfigure}
     \hfill
     \begin{subfigure}{0.32\textwidth}
         \centering
         \includegraphics[width=\textwidth]{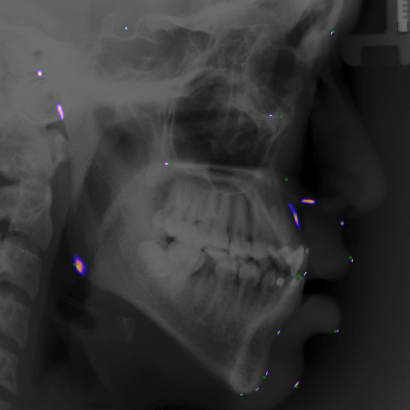}
         \caption{1.72mm MRE}
         \label{fig:poor_sample}
     \end{subfigure}
        \caption{Comparative analysis of heatmaps with green ground truths overlaid on the context image: \textbf{(a)} High-quality heatmap, \textbf{(b)} Novel Salt and Pepper Heatmap generated with $\mathcal{N}(0,\mathbf{I})$ prior, \textbf{(c)}  Medium-quality heatmap with missed prediction around anterior nasal spine landmark.}
        \label{fig:Example_heatmaps}
\end{figure}

Lastly, to meaningfully enhance the salt and pepper activations as seen in \cref{fig:zoomed_x0}, we merge and flatten the activations, by applying a gradually reducing Gaussian blur during the reverse process, such that the model output is convolved by a fixed $13\times13$ kernel with a linearly correlating standard deviation with respect to the timestep. During tuning, a higher $\sigma_\text{max blur}$ performs better which plateaued at approximately $\sigma_\text{max blur}=14$:
\begin{equation}
 \sigma_\text{blur}(t) = \sigma_\text{min blur}+t*\frac{(\sigma_\text{max blur}-\sigma_\text{min blur})}{T-1}.   
\end{equation}

\subsection{Architecture}

Due to domain translation and the availability of relevant architectures, we train a time-encoded U-Net~\cite{dhariwal2021diffusion} from scratch. This is based on PixelCNN~\cite{salimans2017pixelcnn++} and Wide ResNet~\cite{zagoruyko2016wide}, with transformer sinusoidal positional embedding~\cite{vaswani2017attention} to encode the timestep. To improve the low-level feature extraction, we adapt it to have asymmetric encoding-decoding channels with down/up sampling between layers. These were implemented as follows:  (32, 64, 128, 256, 256, 512) (top to bottom) encoding channels, and (256, 128, 64, 64, 32, 32) decoding channels (bottom to top), with QKV spatial self-attention utilised at resolutions 4 and 8 to attend to high-detail features. Finally, we swap the final SiLU activation~\cite{ramachandran2017searching} to a Tanh activation to improve the numerical stability when calculating the log likelihood.

\section{Experiments}
\subsection{Dataset and Augmentations}
We conduct experiments on the publicly available cephalometric dataset~\cite{wang2016benchmark} from the 2015 ISBI Grand Challenge. 
It contains 400 X-rays, split into 150 training, 150 Test 1 (validation), and 100 Test 2 (testing) images. Image resolution is $1935{\times}2400$ with each pixel declared as $0.1\text{mm}{\times}0.1\text{mm}$. Each image contains two ground truth annotations: a junior and senior doctor labelled 19 clinical landmarks. For direct comparison with other works, we downsample the dataset to $640\times800$ pixels and take the mean of the two annotations.

To improve generalisability, we apply several augmentations implemented using the \texttt{imgaug} library (unless otherwise stated). These include: a rotation of $\pm3^{\circ}$, translations of $\pm 10 $ pixels, scale adjustments between $0.95$ and $1.05$, shearing of $\pm 10^\circ$, value multiplication by $\pm 0.5$, elastic transformations with $\alpha{=}500$ and $\sigma{=}30$, a single small regional cutout with random size between 0 and 0.3 of the image size, a random gamma contrast adjustment between 0.5 and 2, and lastly, an initial channel dropout of 0.05\% during the calculation of \cref{eq:q_sample}. Such values were heavily influenced from McCouat \etal.~\cite{mccouat2022contour}.

\subsection{Training}

We train a diffusion U-Net using the combined likelihood loss in \cref{eq:loss} over 120 epochs, taking approximately 7 hours on a V100 GPU with 32GB of VRAM. There is a longer convergence with diffusion due to the artificial augmentation of the training data with respect to $T$. At inference time, the model gradually generates a probability heatmap over the most probable landmark positions as $t=T,...,0$ where we set $T$ as 200. The network is also optimised using the weight-decay AdamW optimiser~\cite{loshchilov2017decoupled} with learning rate $10^{-4}$, weight decay $10^{-4}$, $\beta_1, \beta_2{=}[0.9, 0.999]$ and a batch size of 1.

\subsection{Metrics}

Following the Grand Challenge~\cite{wang2016benchmark}, we use the same two metrics to evaluate our model: \ac{MRE} and \ac{SDR}. \\
\textbf{\ac{MRE}.} Also referred to as the Euclidean distance is defined as the average spatial distance between the ground truth and predicted points. Formally defined as: $\sum_{i=0}^\mathbf{N}(\sqrt{\Delta x_i^2 + \Delta y_i^2})/\mathbf{N}$ (the average $L^2$-norm over the distance vectors).\\
\textbf{\ac{SDR}.} The successful detection of a given landmark point is a predicted landmark within the radius of the ground truth landmark. The typical clinically accepted range is within 2mm~\cite{mccouat2022contour}. Other distances (2mm, 2.5mm, 4mm) are used. When $z$ is the radius to consider a prediction ``successful'' and $N_{all}$ is the set of all landmarks, SDR is formally calculated with $$\frac{|\{\text{MRE}(j)<z:j \in N_{all}\}|}{{|N_{all}|}}\times 100\%.$$ 

\section{Results \& Discussion}

\begin{table}[t]


\begin{tabular}{|l|cccc|cccc|}

\hline
\multicolumn{1}{|c|}{\multirow{3}{*}{Model}} & 
\multicolumn{4}{|c}{Test Set 1} & 
\multicolumn{4}{|c|}{Test Set 2} \\
&
MRE $\downarrow$ & \multicolumn{3}{c|}{SDR(\%) $\uparrow$} & 
MRE $\downarrow$ & \multicolumn{3}{c|}{SDR(\%) $\uparrow$} \\ 
\multicolumn{1}{|c|}{ } & 
(mm) & 2mm & 2.5mm & 4mm & 
(mm) & 2mm & 2.5mm & 4mm \\ 
\hline

Lindner \etal.~\cite{lindner2015fully}     & $1.67 \pm 1.48$ & 74.95 & 80.28 & 89.68 & $1.92 \pm 1.24$ & 66.11 & 72.00 & 87.42 \\

Di Via \etal.~\cite{di2024domain}                      & - & - & - & - & $1.50 \pm \phantom{1.}?\phantom{0}$ & 77.79 & 85.33 & 96.48\\
Chen \etal.~\cite{chen2019cephalometric}   & $1.17 \pm 1.19$ & \pmb{86.67} & \pmb{92.67} & 98.53 & $1.48 \pm 0.77$ & 75.05 & 82.84 & 95.05 \\
Ye \etal.~\cite{ye2023uncertainty}         & $1.16 \pm \phantom{1.}?\phantom{0}$ & 86.25 & 92.18 & \pmb{98.59} & $1.48 \pm \phantom{1.}?\phantom{0}$ & 74.26 & 82.11 & \pmb{95.21} \\
McCouat \etal.~\cite{mccouat2022contour}   & $1.20 \pm \phantom{1.}?\phantom{0}$ & 83.47 & 89.16 & 96.49 & $1.46 \pm \phantom{1.}?\phantom{0}$ & 74.64 & \pmb{83.58} & 93.79 \\ 
Zhong \etal.~\cite{zhong2019attention}     & $1.12 \pm 0.88 $ & 84.91 & 91.82 & 97.90 & $1.42 \pm 0.84$ & \pmb{76.00} & 82.90 & 94.32 \\ \hline
Baseline (No Diffusion)  & $1.27\pm 2.41$ & 82.60 & 88.25 & 96.00 & $1.60 \pm 3.47$ & 74.11 & 81.11 & 93.95 \\
Ours (single-step -- $x_\theta$)       & $1.45 \pm 3.07$ & 80.21 & 86.95 & 95.97 & $1.92 \pm 4.60$ & 69.42 & 78.11 & 91.32 \\
Ours (multi-step)        & $\pmb{1.11 \pm 1.04}$ & 86.04 & 90.84 & 97.37 & \pmb{$1.40 \pm 1.38$} & 75.32 & 82.53 & 94.58 \\
\hline      

\end{tabular}
\vspace{0.5em}
\caption{Localisation results highlighting MRE and SDRs of our method in comparison to existing methods sorted by Test Set 2 MRE. We list a baseline model with the same U-Net architecture as our diffusion method, alongside single and multi-step diffusion results. Our multi-step results show high-precision (MRE) predictions with comparable efficiency (SDR) to other work. While the majority of these use $640{\times}800$,~\cite{di2024domain} train on $512{\times}512$ images thereby getting higher SDR. }
\label{tab:Results}
\end{table}

We compare our results against original challenge entries~\cite{Ibragimov2014AutomaticCX,lindner2015fully} and current representative state-of-the-art~\cite{chen2019cephalometric,ye2023uncertainty,mccouat2022contour,zhong2019attention}. The experiments shown in \cref{tab:Results} highlight an overall highest precision \ac{MRE} for our multi-step diffusion model, and comparable \acp{SDR} across all radii. We also highlight a baseline model, which is the same U-Net we use for diffusion, trained with the same NLL loss as \cref{eq:nll-loss}, augmentations and optimiser. This model highlights the improvement of using diffusion as a spatial probability generator. Further, the single-step generation - predicting $x_0$ from $x_T$ - massively under-performs multi-step diffusion, highlighting that the subsequent predictions do aid and improve the final prediction by accumulating around the correct position.

Chen \etal.~\cite{chen2019cephalometric} and Ye \etal.~\cite{ye2023uncertainty} have the most efficient predictions due to attention pyramidal fusion modules. The majority of the top performing models utilise some form of ensembling or multi-level feature fusion. 

Upon analysing our measured metrics per landmark, it emerges that landmark 16, the soft tissue pogonion~\cite{lindner2015fully}, has an incredibly high inter-annotator variability on the second test set. When taking the average of the two annotations as ground truth it results in misleading numerical results: the prediction may have given a perfectly valid output according to one annotator but may be a few millimetres from the second annotator. In the future, it may be more relevant to also measure the ratio between the two ground truth annotations and the total $L^2$ distance between the predicted point and the two annotations. This would work best for annotations with high variability that lie on an edge. 

By way of an ablation experiment, we explored the impact of adjusting $T$ on the MRE and SDR and found that the multi-step performance hit an MRE of 1.50mm when $T=50$, and continued  gradually to decrease and plateau at 200 with minor random fluctuations expected with diffusion models. Further, when training with Ho \etal.'s \cite{ho2020denoising} primary formulation of $\epsilon_\theta$, the model struggles to distinguish how input channels relate to specific output channels, and leads to meaningful heatmaps but the highest activation for a given channel would lie over another channel's landmark. Increasing T slightly improves this but ultimately never consistently performed better than 4mm MRE.



\section{Conclusion}


Multi-step diffusion for automatic  landmark detection has been shown to be effective. Optimising $x_\theta$ (over $\epsilon_\theta$) brings significant training time reductions; additionally the iterative generation brings more scope to reduce attention mechanisms in future work, reducing the memory footprint. 
In the future, we will explore alternative formulations, such as predicting regions rather than single-pixels to improve the efficiency. Moreover, the diffusion process may perform even better if the diffusion distribution at $x_T$ was instead an informed prior leading the output to hone in on an efficient and accurate prediction. In addition to using an updated diffusion formulation with improved inference speeds.


\section{Acknowledgements}


This work was funded by an EPSRC 
Research Scholarship [XXXX-YYYY]. Human subject data was available publicly.
The authors would like to acknowledge the use of 
the University of Oxford Advanced Research Computing (ARC) in carrying out this work. \footnote{\url{http://dx.doi.org/10.5281/zenodo.22558}}

\bibliographystyle{splncs04}
\bibliography{citations}
%





\end{document}